\documentclass[10pt, a4paper]{article}

\pdfoutput=1

\usepackage{lrec}
\usepackage{graphicx}
\usepackage{tabularx}
\usepackage{soul}
\usepackage{epstopdf}
\usepackage[T1]{fontenc}
\usepackage[utf8]{inputenc}

\usepackage[hyphens]{url}
\usepackage{hyperref}
\usepackage{xstring}
\usepackage{color}

\newcommand{\norex}[1]{\textit{#1}}
\newcommand{\eng}[1]{`#1'}

\usepackage{array}
\newcolumntype{P}[1]{>{\raggedright\arraybackslash}p{#1}}

\newcommand{\DEVELOPMENT}{1} % 1= color-highlight revisions, 0=no highlighting
\usepackage{ifthen}
\ifthenelse{\DEVELOPMENT = 0}{
	\newcommand{\rev}[1]{\textcolor{red}{#1}}	
}{
	\newcommand{\rev}[1]{\textcolor{black}{#1}}	
}

\title{Building a Norwegian Lexical Resource for Medical Entity Recognition}

\name{Ildikó Pilán\textsuperscript{*}, Pål H. Brekke\textsuperscript{$\dagger$}, Lilja Øvrelid\textsuperscript{*}}

\address{\textsuperscript{*}Department of Informatics, University of Oslo, \textsuperscript{$\dagger$}Department of Cardiology, Oslo University Hospital Rikshospitalet
  \\
         Oslo, Norway \\
         ildikop@ifi.uio.no, pabrek@ous-hf.no, liljao@ifi.uio.no\\}

\abstract{We present a large Norwegian lexical resource of categorized medical terms. The resource merges information from large medical databases, and contains over \rev{77,000} unique entries, including automatically mapped terms from a Norwegian medical dictionary. We describe the methodology behind this automatic dictionary entry mapping based on keywords and suffixes and further present the results of a manual evaluation performed on a subset by a domain expert. The evaluation indicated that ca. 80\% of the mappings were correct. 
\\ \newline  }

\begin{document}

\maketitleabstract

\section{Introduction}

Named Entity Recognition (NER) is a common task within the area of clinical Natural Language Processing (NLP) with the aim of extracting critical information such as diseases and treatments from unstructured texts \cite{Fri:Ald:Aus:1994,Xu:Ste:Doa:2010,Jag:Hon:2016}.

Current neural approaches to NER typically require a large amount of annotated data for a reliable performance \cite{Ma:Hovy:2016,Lam:Bal:Sub:2016}. Distant supervision \cite{Min:Bil:Sno:2009}, however, relaxes this constraint on the training data size thanks to the combined use of information from lexical resources, a small amount of training data and large amounts of raw data. This technique has been successfully applied also in the biomedical and clinical domain \cite{Fri:Wu:Rat:2017,Sha:Liu:Ren:2018}.
In absence of even a small amount of annotated data, categorized lexical resources can also be used as gazetteers in rule-based approaches.

There is currently no large and freely available lexical resource with categorized entity types for Norwegian medical terms to be used for clinical NER with distant supervision.
This paper presents an effort to  create such a resource by collecting and merging lists of terms available from a number of other smaller and more specialized resources. We implement and describe an automatic mapping method which is applied to a dictionary containing a variety of definitions for relevant terms and present an evaluation of this mapping using both inter-resource overlap and manual evaluation performed by a domain expert. The resulting lexical resource will be made freely available.

\section{Background}

Medical Entity Recognition often makes use of lexical resources such as lists of disease names derived from the International Statistical Classification of Diseases and Related Health Problems (ICD) resource \cite{world2004icd} or from disease information from general resources, such as the Medical Subject Headings  \cite[MeSH]{lipscomb2000medical}.
There has been quite a bit of work aimed at creating semantic lexicons for use in NLP from such domain-specific resources \cite{Joh:1999,Liu:Wu:Din:2012}.

Automated extraction of medical entities from clinical text has been the topic of several research efforts more recently, a majority aimed at English \cite{Xu:Ste:Doa:2010,Jag:Hon:2016} and Chinese clinical text \cite{Wu:Min:Xu:2017}. For a language that is very closely related to Norwegian, \newcite{Ske:Kvi:Nil:2014} developed and evaluated an entity detection system for Findings, Disorders and Body Parts in Swedish.
In order to alleviate the need for manual annotation, distant supervision has recently been applied also to entity recognition in the medical domain for English and Chinese  \cite{Sha:Liu:Ren:2018,nooralahzadeh-etal-2019-reinforcement}.

\section{Norwegian Medical Terminology Resources}
\label{sec:resources}

There are a number of resources which contain Norwegian medical terms that could in principle be relevant for NER. 

The \norex{Medisinsk ordbok} (MO) \eng{Medical Dictionary} \cite{medisinsk-ordbok} contains 23,863 Norwegian medical terms of various kinds including, among others, names of diseases and treatments, anatomical terminology as well as types of medical specialists and specialization areas. The dictionary contains synonyms and one or more definitions of these terms depending on the number of senses per entry.  

\rev{O}ther rich sources of Norwegian medical terms and their corresponding standardized codes are available from the website of \norex{Directoratet for e-helse} \eng{Norwegian Directorate for e-health}. One is the Norwegian equivalent of the 10th Revision of ICD (ICD-10). The widely-used resource lists both coarse and fine-grained codes and corresponding terms relative to diseases, symptoms and findings. \rev{An}other source is the Procedure Coding Schemes list (referred to as PROC here), which includes diagnostic, medical and surgical intervention names and codes \cite{prosedyrekod}.
\rev{Moreover, Laboratoriekodeverket\footnote{\url{https://ehelse.no/kodeverk/laboratoriekodeverket}} \eng{List of laboratory codes} (LABV) contains various substance names relevant in laboratory analyses. The web page of this list also includes a shorter list of anatomical locations, which we refer to as ALOC here.
Yet another resource available from the Directorate's web site is the Norwegian equivalent of the International Classification of Primary Care (ICPC-2), which includes diagnosis terms as well as health problem and medical procedure names.} 

The FEST (\norex{Forskrivnings- og ekspedisjonsstøtte}, \eng{Prescribing and dispensing support}) database\footnote{\url{https://legemiddelverket.no/andre-temaer/fest}} contains information about all medicines and other goods that can be prescribed in Norway. FEST is a publicly available resource published by \norex{Statens legemiddelverk} \eng{The Norwegian Medicines Agency}.

\newcite{Ram:Bre:Nyt:2018} present a corpus of synthetically produced clinical statements about family history in Norwegian (here dubbed FAM-HIST). The corpus is annotated with clinical entities relating to family history, such as Family Member, Condition and Event, as well as relations between these.

\section{Automatic Dictionary Entry Mapping Method}
\label{mapping_method}
The use of dictionary definitions as a source of semantic information has been the topic of quite a bit of research in lexical semantics, from the early work of \newcite{Mar:Ahl:Eve:1986} where patterns in the dictionary definitions along with suffix information gave rise to a semantic lexicon to more recent efforts to embed dictionary definitions in order to derive semantic categories for phrasal units \cite{Hil:Cho:Kor:2016}.

In this work, we map entries from the MO dictionary to categories, i.e.~to medical entity types. We identify 12 different types of entity categories based on previous work \cite{10.1016/j.jbi.2013.08.004} and the inspection of MO entries. We then implement a rule-based mapping method relying on suffixes and keywords. 

\subsection{Mapping Strategies}

The mapping method consists of four different mapping strategies: two relying on the entries themselves and two deriving the mapped category from the definitions. One of these is suffix based, the others operate based on keywords. In what follows, we describe each of these strategies in detail.

\paragraph{Suffix-based mapping} (\textit{strategy SUFF}) This strategy consists of mapping an entry to a category whenever its last characters match a specific suffix. Many medical terms have Greek or Latin origin resulting in suffixes that give rather clear indications of the category of an entry. We compile a list of suffixes based on both frequently occurring suffixes in the data and an online resource\footnote{\url{https://en.wikipedia.org/wiki/List_of_medical_roots,_suffixes_and_prefixes}}. We only include suffixes and endings which can be mapped to an unambiguous category in the majority of cases. The complete list used for the mapping is presented in Table \ref{tbl:suffixes}. 

\begin{table}[h]
\begin{center}
\begin{tabular}{|p{2cm}|p{5.2cm}|}
\hline
\textbf{Category} & \textbf{Suffixes}   \\
\hline

CONDITION & -agi, -algi, -algia, -blastom, -cele, 
\newline-cytose, -donti, -dynia,
-emi, -emia, 
\newline-epsi, -ism, -isme, -ismus, -itis, -oma, -pati, -plasi, -plegi, -ruptur, -sarkom, -sis, 
-trofi, -temi, -toni, -tropi \\
DISCIPLINE & -iatri, -logi \\
MICROORG & -coccus, -bacillus, -bacter \\
PERSON & -iater, -olog \\
PROCEDURE & -biopsi, -grafi, -metri, -skopi, -tomi \\
SUBSTANCE & -cillin \\
TOOL & -graf, -meter, -skop \\
\hline
\end{tabular}
\caption{\label{tbl:suffixes} Suffix mapping.}
\end{center}
\end{table}

\paragraph{Keyword-based mapping} Mapping entries to keywords is primarily used to map an entry to a category based on the first noun occurring in their definition (\textit{strategy KW-1N}). To be able to detect first nouns, definitions are tokenized and part-of-speech tagged with UDPipe \cite{straka2016udpipe}.

To create a list of keywords for the mapping, we inspect the 200 most frequent nouns in the definitions and manually map the ones with a strong indication of a single category. We complement this with other frequent nouns which can be good indicators of a category. This results in a list of 168 mapped keywords, see Table \ref{tbl:cats} for some examples.

\begin{table*}[t]
\begin{center}
\begin{tabular}{|l|P{3.5cm}|P{4.7cm}|P{4.5cm}|}
\hline
\textbf{Category}    & \textbf{Description} &  \textbf{Example keywords} &   \textbf{Mapped entry examples} \\
\hline
ABBREV 	& abbreviations, acronyms & \norex{forkortelse} \eng{abbreviation} & \textit{Ahus}, \textit{ADH}\\
ANAT-LOC 	& anatomical locations & \norex{celler} \eng{cells}, \norex{muskel} \eng{muscle}, \norex{kroppsdel} \eng{bodypart} & \norex{fødselskanalen} \eng{birth-channel}, \norex{halsmusklene} \eng{throat-muscles} \\ 
CONDITION 	& diseases, findings & \norex{sykdom} \eng{disease}, \norex{tilstand} \eng{condition}, \norex{mangel} \eng{deficiency}  & \textit{leukemi} \eng{leukemia}, \textit{leverkoma} \eng{hepatic coma} \\
DISCIPLINE 	& medical disciplines & \norex{studium} \eng{study}, \norex{forskning} \eng{research}, \norex{teori} \eng{theory} & \norex{dietetikk} \eng{diethetics},  \norex{biomekanikk} \eng{biomechanics} \\
MICROORG 	& microorganisms of different kind & \norex{bakterie} \eng{bacteoria}, \norex{organisme} \eng{organism}, \norex{virus} \eng{virus} & \norex{kolibakterie} \eng{colibacteria}, \norex{blodparasitter} \eng{blood parasites} \\ 
ORGANIZATION & institutions and organizations & \norex{foretak} \eng{company}, \norex{institutt} \eng{institute} & \norex{Røde Kors} \eng{Red Cross}, \norex{sanatorium} \eng{sanatorium}	\\  
PERSON 		& types of practitioner or patient & \norex{lege} \eng{doctor}, \norex{pasient} \eng{patient}, \norex{individ} \eng{individual} & \norex{myop} \eng{myope}, \norex{nevrolog} \eng{neurologist} \\ 
PHYSIOLOGY 	& physiological functions & \norex{refleks} \eng{reflex}, \norex{sammentrekning} \eng{contraction} & \norex{adsorpsjon} \eng{absorption}, \norex{forbrenning} \eng{burning} \\
PROCEDURE 	& procedure and treatment types & \norex{behandling} \eng{treatment}, \norex{fjerning} \eng{removal} & \norex{nyrebiopsi} \eng{kidney biopsy}, \norex{detoksifisering} \eng{detoxification} \\ 
SERVICE & types of services & \norex{tjeneste} \eng{service}, \norex{omsorg} \eng{care} & \norex{tannhelsetjeneste}, \eng{
dental service}, \norex{sjelesorg} \eng{counseling} \\
SUBSTANCE 	& medicines and other substances & \norex{stoff} \eng{substance}, \norex{løsning} \eng{solution}‚  \norex{medikament} \eng{drug} & \norex{aspartam} \eng{aspartam}, \norex{paracetamol} \eng{paracetamol}\\
TOOL 		& instruments and tools & \norex{instrument} \eng{instrument}, \norex{verktøy} \eng{tool} & \norex{diatermikniv} \eng{diathermy blade}, \norex{defibrillator} \eng{defibrillator}\\
\hline
\end{tabular}
\caption{\label{tbl:cats} List of entity type categories, keywords and mapped entries.}
\end{center}
\end{table*}

When mapping, we require the first noun of a definition to either (i) exactly match a keyword or (ii) to contain it. The latter is only applied for keywords longer than 4 characters to avoid short sequences which might over-generate false positives (e.g. \norex{tap} \eng{loss} for \norex{katapleksi} \eng{cataplexy}).

When checking for contained keyword, we limit the position of the keyword match to the second character onward in the first noun to approximate the occurrence of a keyword as the second part of a compound as this is more indicative of categories. Given that many dictionary entries are also compounds, we apply the mapping based on contained keyword also to the entries themselves (\textit{strategy KW-E}). 

When applying keyword-based mapping to definitions, before detecting the first noun, we remove those nouns and phrases which have little added semantic value relevant for the category. These include prepositional phrases forming a complex noun phrase typical of definitions (e.g.~\norex{form av} \eng{form of}), nouns not indicative of a category (e.g. \norex{uttrykk} \eng{expression}) and abbreviations (\norex{plur.} \eng{plural}, \norex{lat.} \eng{Latin}).

During the mapping procedure, first each strategy casts a vote on the category. In case of multiple votes with a disagreement, the category is based on a single mapping strategy chosen following a specific order, starting from the strategy with the highest expected precision and continuing with the ones with increasingly high recall as follows: SUFF $\rightarrow$ KW-E $\rightarrow$ KW-1N.
After a first iteration of mapping, we perform a second iteration and map uncategorized entries if there is an entry already mapped available for the first noun in their definition (\textit{strategy ITER}).

The MO resource contains altogether 2,387 synonyms, which were treated as separate entries with the same definition. The number of entries with multiple meanings (and definitions) were merely 360 in total, amounting to 1.5\%. Since such polysemous entries were so rare, we consider only the first sense of each entry.

\rev{The methodology outlined above could be applied also for categorizing medical terminology in other languages via, for example, machine translating the list of keywords (or terms) used and making small language-specific orthographic adjustments to the suffix mappings from Table \ref{tbl:suffixes}. Such suffixes are often adapted to the orthographic conventions of a certain language, as also \newcite{grigonyte2016swedification} found in the case of Swedish.}

\section{Mapping Results}

The results of the category mapping for MO based on the methodology outlined in Section \ref{mapping_method} is presented in Table \ref{tbl:res-mapping}. 

\begin{table}[h]
\begin{center}
\begin{tabular}{|l|c|}
\hline
\textbf{Category} & \# \textbf{entries} \\
\hline
CONDITION &	5,522\\
SUBSTANCE &	2,216\\
PROCEDURE & 1,467\\
DISCIPLINE &	418\\
ANAT-LOC &	408\\
PERSON 	&	282\\
MICROORG &	227\\
ABBREV &	216\\
TOOL 	&	210\\
PHYSIOLOGY & 132\\
ORGANIZATION & 81\\
SERVICE &	48\\
\hline
\textbf{Total mapped} & \textbf{11,227} \\
\hline
Not mapped	&	12,636\\
\hline
\textbf{Total} &  \textbf{23,863}\\
    \hline
\end{tabular}
\caption{\label{tbl:res-mapping} Mapping results for MO.}
\end{center}
\end{table}

The percentage of mapped entries was 47\%, almost half of all available entries in MO. \rev{The other terms, which were not mapped, did not match either any of the suffixes or the keywords used. The latter includes, among others, cases where the first noun in the definition was a synonym of the term and hence too specific to be included in the list of keywords used (e.g.~the term \norex{klorose} \eng{chlorosis}, a type of anemia occurring mosty in adolescent girls is defined using \norex{jomfrusyk} \eng{virgin sick}).} 

\rev{Based on some manual inspection, most non-mapped terms would fit one of the categories proposed, with few exceptions that might lead to rather small categories, such as regulations (e.g.~\norex{internasjonalt helsereglement} \eng{international health regulations}).} 

Several non-mapped terms should belong to the ANAT-LOC category. The proposed keyword-based methods would often be ambiguous for these terms and could indicate either an anatomical location or a medical condition related to it. For example, both the ANAT-LOC \norex{hjertekammer} \eng{ventricle} and the CONDITION \norex{panserhjerte} \eng{armoured heart} contain the keyword \norex{hjerte} \eng{heart} and have this word also as the first noun in their definition, their category could thus not be determined by our method. Additional databases containing a detailed list of anatomical location terms are therefore particularly useful for expanding our resource. 

We also inspected the distribution of the mapping strategies used (see Table \ref{tbl:res-strat}), where MULTI stands for a category selected based on the unanimous vote of multiple voting strategies. 
We can observe that the most frequently used strategy was KW-IN. Mappings based on multiple voting strategies selecting the same category were also rather common, occurring in 21\% of all mapped entries.  

\begin{table}[h]
\begin{center}
\begin{tabular}{|l|c|}
\hline
\textbf{Strategy} & \# \textbf{entries} \\
\hline
KW-1N 	 & 5,489 \\
MULTI 	 & 2,397 \\
ITER 	 & 1,157 \\
SUFF 	 & 1,096 \\
KW-E 	 & 1,088 \\
\hline
\textbf{Total}    & \textbf{11,227} \\
\hline
\end{tabular}
\caption{\label{tbl:res-strat} Distribution of mapping strategy use.}
\end{center}
\end{table}

\section{Resource Merging}

The mapped MO entries were complemented with data from the other resources described in Section \ref{sec:resources} The mapping for these resources was straightforward since each resource contained either one specific type of entity or manual annotation was available.

\rev{At a closer inspection, we found that the ALOC list contains, besides anatomic locations, several terms which could belong to more than one category depending on the context of their use, e.g. \norex{tracheostomi} \eng{tracheostomy} could either be ANAT-LOC referring to the hole created during a tracheostomy or it could refer to the procedure itself. These cases were mapped to PROC for reasons of consistency with the suffix-based mapping applied, but it might be worth to accommodate multiple categories in future versions. This list has been manually revised by a medical expert who disambiguated the category consistently with the mapping methodology used.}

From FAM-HIST, we collected all occurrences of condition and event entities and mapped them to our CONDITION category. 
The SUBSTANCE category was augmented, in part, based on the FEST resource. The terms collected from FEST included substance names (also in English, when available) as well as medical product names with and without strength information.
From ICD-10, both the disease names corresponding to the 3 and the 4 digit codes were preserved. Only 16\% of the ICD codes were 3 digit codes.

\rev{From ICPC-2, we included all terms, sub-terms and short forms under the CONDITION category except for the terms appearing in the \textit{Procedure codes} chapter, which were mapped to the PROCEDURE category. Terms from the \textit{Social problems} chapter were excluded as most of these were not strictly speaking medical conditions (e.g. \norex{lav inntekt} \eng{low income}). We observed a minor difference compared to ICD between some terms associated to the same code (e.g.~\norex{Blindtarmsbetennelse} vs.~\norex{Uspesifisert appendisitt} for code K37, appendicitis).}

\rev{In the case of LABV, we included under the SUBSTANCE category all substance names, medicine and other medical product names and brands together with type and strength information when available (e.g.~\norex{Kortison Tab 25 mg} \eng{Cortisone Tablet 25 mg}).}
\rev{Lastly, all codes from PROC were included without any filtering.}

Table \ref{tbl:res-exp} presents the amount of total entries available from various resources compared to MO. The total number of categorized entries created after merging \rev{and excluding all inter-resource overlaps was 78,105 with the original casing and 77,320 when normalizing all entries to lowercase}.

\begin{table}[h]
\begin{center}
\begin{tabular}{|l|c|c|}
\hline
\textbf{Resource}	& \textbf{Category} &	\# \textbf{entries} \\
\hline
MO & Multiple & 11,227 \\
\hline
\rev{ALOC} & \rev{Multiple} & \rev{287} \\
FAM-HIST & COND. & 283 \\
FEST & SUBST. & 26,234 \\
ICD-10	& COND. & 10,765 \\
\rev{ICPC-2} & \rev{Multiple} & \rev{9,420} \\
\rev{LABV} & \rev{SUBST.} & \rev{14,193} \\
PROC & PROC. & 8,883\\
\hline
\textbf{Total} & N/A & \rev{\textbf{81,292}} \\
\hline
\end{tabular}
\caption{\label{tbl:res-exp} Number and type of entries in different resources.}
\end{center}
\end{table}

\section{Resource-based Automatic Evaluation}

Thanks to a certain amount of overlap between the mapped MO entries and the other resources, we can use information from the latter to automatically evaluate the former. Table \ref{tbl:res-autom-eval} shows the overlap and the percentage of correct mappings.

\begin{table}[ht]
\begin{center}
\begin{tabular}{|l|c|c|c|}
\hline
\textbf{Resource}	&	\# \textbf{overlap} & \textbf{Correct (\%)} & \textbf{Category}\\
\hline
\rev{ALOC} & \rev{33} & \rev{57.6} & \rev{Multiple} \\
FAM-HIST &	22 &	63.6	&	COND. \\
FEST & 744 & 97.3 & SUBST. \\
ICD-10  	&	307	&	97.7	&	COND. \\
\rev{ICPC-2} & \rev{886} & \rev{94.0} & \rev{Multiple}\\
\rev{LABV} & \rev{297} & \rev{85.5} & \rev{SUBST.}\\
PROC 	&	89 &	97.8 &	PROC. \\
\hline
\end{tabular}
\caption{\label{tbl:res-autom-eval} Evaluation results of the mapped MO entries.}
\end{center}
\end{table}

On average, \rev{85\%}  mappings were correct \rev{out of the total of 2,378 overlapping terms from the resources listed in Table \ref{tbl:res-autom-eval}. Approximately 21\% of all mapped terms from MO were thus evaluated (and corrected) automatically with the help of the other resources.
Most misclassifications occurred with the ALOC and FAM-HIST resources and concerned the ANAT-LOC and CONDITION categories.} 

\section{Manual Evaluation}

Given that the overlap between MO and the other resources was limited to certain categories, we further performed a manual evaluation of the automatically mapped MO entries in order to assess their quality.

We randomly selected 1,128 terms to evaluate manually, aiming at a balanced amount per category (100 each) and mapping method. We included all available terms for categories where the total amount of terms remained below 100. The terms were categorized by a medical expert without access to the automatically mapped categories and the mapping method used. We present the per-category precision and recall in Table \ref{tbl:res-man-eval}, where the number of terms in the last column refers to manually assigned labels.

\begin{table}[ht]
\begin{center}
\begin{tabular}{|l|c|c|c|}
\hline
\textbf{Category}	&	\textbf{Prec} & \textbf{Recall} & \textbf{\#}\\
\hline
ABBREV         & 0.969   & 0.750   & 124     \\
ANAT-LOC       & 0.928   & 0.796   & 113     \\
CONDITION      & 0.915   & 0.623   & 138     \\
DISCIPLINE     & 0.702   & 0.855   & 69      \\
MICROORG       & 0.871   & 0.976   & 83      \\
ORGANIZATION   & 0.548   & 0.714   & 56      \\
PERSON         & 0.593   & 0.923   & 52      \\
PHYSIOLOGY     & 0.710   & 0.815   & 81      \\
PROCEDURE      & 0.793   & 0.821   & 84      \\
SERVICE        & 0.667   & 0.468   & 47      \\
SUBSTANCE      & 0.809   & 0.905   & 84      \\
TOOL           & 0.846   & 0.906   & 85      \\
\hline
\textbf{Total} & \textbf{0.779}  & \textbf{0.796}   & \textbf{1,016}    \\
\hline
ORG+SER   & 0.830   & 0.854   & 103     \\
\hline
\textbf{Total} ORG+SER & \textbf{0.815}  & \textbf{0.839}   & \textbf{1,016}    \\ 
\hline
\end{tabular}
\caption{\label{tbl:res-man-eval} Manual evaluation results.}
\end{center}
\end{table}

112 terms were labeled as `OTHER' in cases where a term did not belong to any of the 12 categories indicated or when terms were outside of the area of expertise of the evaluator. Table \ref{tbl:res-man-eval} excludes OTHER, as this was not part of the automatically mapped categories. 
The percentage of correctly categorized entries including and excluding terms labeled as OTHER, was  71.5\% and 79.4\% respectively.
In 20 cases, SERVICE and ORGANIZATION were indicated as alternative labels to each other. We therefore compute evaluation measures also with these two categories merged (ORG+SER). This yields in total 82\% correct labels when excluding OTHER.

\begin{figure}[ht]
\centering
\includegraphics[width=0.48\textwidth,trim={0 0 3.5cm 0},clip]{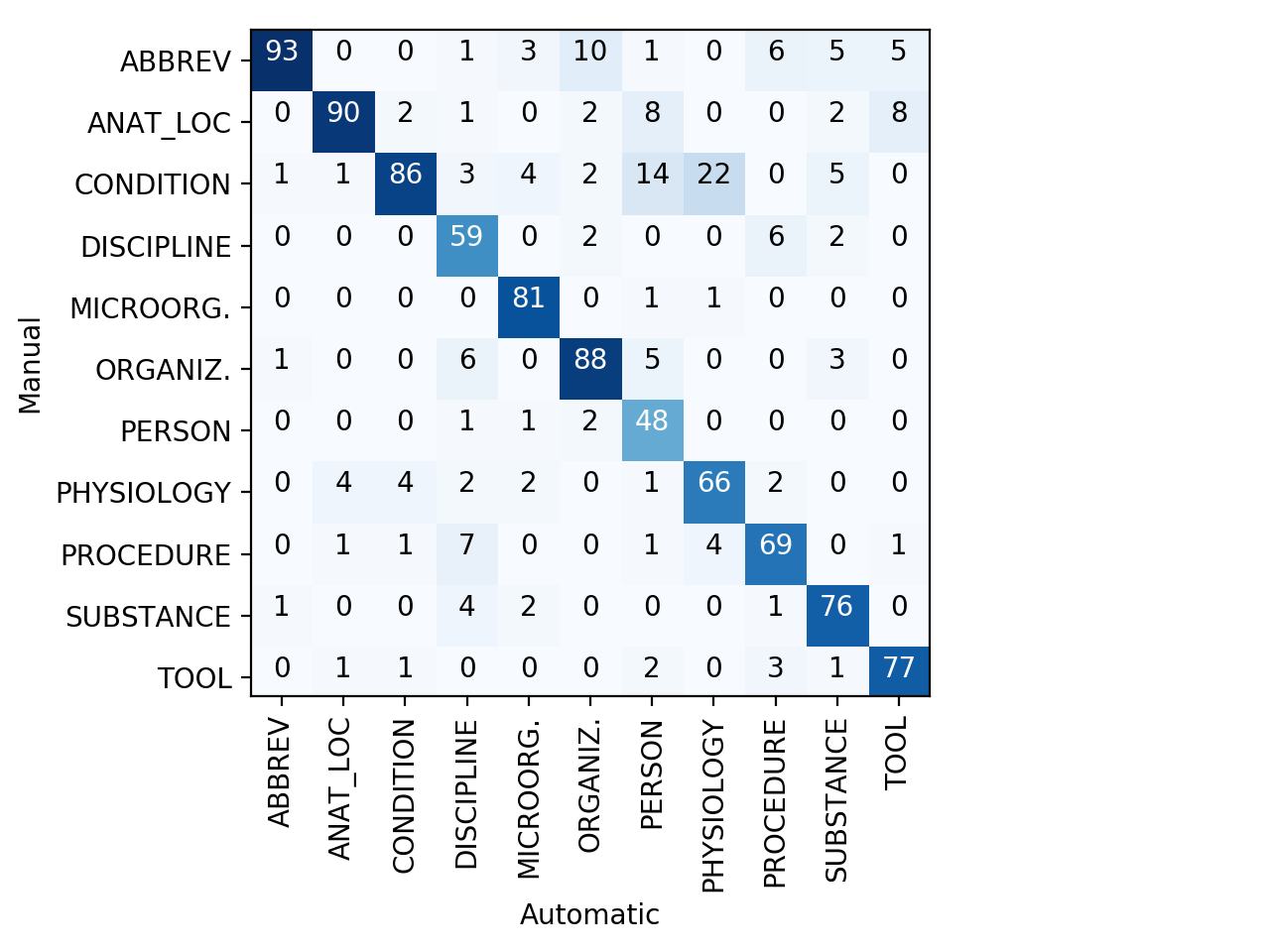}
\caption{Confusion matrix over categories. }
\label{img:cm}
\end{figure}

According to the confusion matrix in Figure \ref{img:cm}, most automatic categorization errors occurred between CONDITION and PHYSIOLOGY. (SERVICE was mapped to ORGANIZATION here.) 

Errors related to the PERSON category were mostly connected to the use of \norex{person} as keyword with the KW-E strategy, which generated false positives such as \norex{schizoid personlighetstype} \eng{schizoid personality type}. 
\rev{Some categorization errors occurred because of the lack of prefix information, e.g.~in the case of the keyword \norex{refleks} \eng{reflex} in \norex{arefleksi} \eng{areflexia} and \norex{hyperrefleksi}  \eng{hyperreflexia}, which were both mapped to PHYSIOLOGY instead of CONDITION. This indicates that taking into consideration prefixes would contribute to improving the automatic categorization, especially for the KW-E strategy.}
\rev{The category label confusions between TOOL and ANAT-LOC originated from the keyword \norex{apparat}, which proved to be ambiguous for the proposed categories, not only meaning \eng{device} and thus mappable to TOOL, but also meaning \eng{apparatus, system} as in \norex{immunapparatet} \eng{immune system} and thus belonging to ANAT-LOC.}

Most correct mappings (88.3\%) with a single strategy were obtained using suffixed (SUFF), followed by the keyword mapping from first nouns (KW-1N, 79.9\%) and entries (KW-E, 76.6\%). The iterative mapping (ITER) yielded considerably fewer correct mappings, only 64.7\%. When multiple strategies opted for the same category label, 98.2\% of terms were correctly categorized.

\rev{As a final step during the resource creation, we revised the automatic categories based on the manually assigned ones. The updated count of terms per category in the resource after merging with other databases (eliminating overlap) and incorporating the evaluation results is reported in Table \ref{tbl:res-final}}.

\begin{table}[h]
\begin{center}
\begin{tabular}{|l|c|}
\hline
\textbf{Category} & \# \textbf{entries} \\
\hline
SUBSTANCE  &  41,365 \\
CONDITION  &  24,071 \\
PROCEDURE  &  10,420 \\
ANAT-LOC  &  658 \\
DISCIPLINE  &  387 \\
ABBREV  &  236 \\
PERSON  &  232 \\
TOOL  &  216 \\
MICROORGANISM  &  193 \\
OTHER  &  112 \\
PHYSIOLOGY  &  112 \\
ORGANIZATION  &  103 \\
\hline
\textbf{Total (original casing)} & \textbf{78,105} \\
\hline
\end{tabular}
\caption{\label{tbl:res-final} \rev{Final term counts per category in the resource.}}
\end{center}
\end{table}

\section{Conclusion}
We introduced the first Norwegian lexical resource of categorized medical entities and provided an overview of the process of its creation. The resource unites information from medical databases as well as entries automatically mapped from a medical lexicon. A manual evaluation of a subset of the mapped terms confirmed that the automatic mappings were of a suitable quality to be used as additional supervision signal with machine learning based NER approaches.
In future work we plan to apply the resource in medical entity recognition for Norwegian, using it to provide initial categories for distant supervision. \rev{We also plan to perform annotations with multiple raters and measure inter-annotator agreement for the proposed categories}.

\section{Acknowledgments}
This work is funded by the Norwegian Research  Council  and more specifically by the BigMed project, an IKTPLUSS Lighthouse project.

\section{Bibliographical References}
\label{main:ref}

%\bibliographystyle{lrec}
%\bibliography{anthology, lrec2020}

\end{document}